\documentclass[acmsmall,screen,nonacm]{acmart}

\usepackage{booktabs}
\usepackage{multirow}

\setcopyright{none}
\acmJournal{POMACS}
\acmYear{2027}

\begin{document}

\title[When Does Risk-Controlled Throughput Forecasting Control Risk?]{When Does Risk-Controlled Throughput Forecasting Actually Control Risk? A Reproduction Study on LEO Satellite Links}

\author{Mayan Sharma}
\email{mayan25sharma@gmail.com}
\affiliation{%
  \department{School of Computer Science and Engineering}
  \institution{Vellore Institute of Technology}
  \city{Chennai}
  \state{Tamil Nadu}
  \country{India}}

\author{Kriti Saini}
\affiliation{%
  \department{School of Computer Science and Engineering}
  \institution{Vellore Institute of Technology}
  \city{Chennai}
  \state{Tamil Nadu}
  \country{India}}

\author{Devansh Behl}
\affiliation{%
  \department{School of Computer Science and Engineering}
  \institution{Vellore Institute of Technology}
  \city{Chennai}
  \state{Tamil Nadu}
  \country{India}}

\author{Suganya R}
\affiliation{%
  \position{Professor}
  \department{School of Computer Science and Engineering}
  \institution{Vellore Institute of Technology}
  \city{Chennai}
  \state{Tamil Nadu}
  \country{India}}

\author{S. Vijayaragavan}
\affiliation{%
  \position{Professor of Practice}
  \institution{Nelco Network Products Ltd.}
  \country{India}}

\authorsaddresses{Authors' Contact Information: Mayan Sharma, Kriti Saini, Devansh Behl and Suganya R, School of Computer Science and Engineering, Vellore Institute of Technology, Chennai, Tamil Nadu, India; S. Vijayaragavan, Nelco Network Products Ltd., India. Corresponding author: Mayan Sharma, \href{mailto:mayan25sharma@gmail.com}{mayan25sharma@gmail.com}.}

\begin{abstract}
Risk-budgeted throughput forecasting has been proposed as a way to make LEO satellite capacity estimates safe for admission control: a point forecast is converted into a conservative bound whose overestimation rate is held at a user-specified budget. We reproduce the state of the art, BG-CFQS, on the three Starlink traces its authors used and on an independent fourth dataset, and report four findings.

First, its risk guarantee is reproducible but conditional on the calibration split. Under a random split of the same data we recover every published quantity, including the risk-pass column at 3 of 3 datasets. Under a contiguous temporal split, which is what a deployed terminal faces, it holds on 1 of 3.

Second, and independent of any split, the method cannot serve a risk budget below 0.15. Its published candidate quantile set is $T = [0.15, 0.40]$, and when the budget is tighter than the risk at the most conservative candidate the selection interval collapses to a point and the achieved rate pins there. On all four datasets the achieved rate is identical at budgets of 0.05, 0.10 and 0.15, overshooting by 1.4x to 3.7x at the tightest. The paper evaluates only at 0.35, where the floor never binds.

Third, we show that static conformal calibration cannot improve a ranking at all, only shift an operating point along an unchanged curve, and confirm the consequence exactly on latency spike detection. Online calibration is not so constrained and improves detection AUPRC by up to 51\%.

Fourth, we evaluate the natural remedy, group-conditional online recalibration, and find that it splits into two mechanisms with very different standing. The online part replicates: it tracks the budget within 1\% at every point from 0.05 to 0.35, where static calibration is 16 to 21\% off, and it dominates static calibration on both tail risk and accuracy at matched achieved risk. The group-conditional part does not: on three of four datasets, conditioning on any covariate we have, including \emph{measured} serving-satellite geometry, is worse than not conditioning at all. We attempted to predict which datasets benefit, from a statistic computable on the calibration split, and failed. We report that failure in full.

What is new here is empirical, not algorithmic. The mechanisms we evaluate are all published, and Section~\ref{sec:prior} says which paper each comes from. We claim the findings, not the methods.
\end{abstract}




\maketitle

\section{Introduction}

A LEO access link is a hard thing to allocate bandwidth over. Capacity moves on a 15-second scheduling cadence~\cite{tanveer2023constellations,casparsen2026latency}, handovers are frequent, and the achievable rate varies by more than a factor of three within a single hour~\cite{liu2025starnet}. An allocator that acts on a point forecast will over-promise whenever the forecast is optimistic, and over-promising drops sessions.

The natural response is to allocate against a \emph{conservative bound} rather than a point forecast, and to control how often that bound is wrong. BG-CFQS~\cite{xie2026bgcfqs} does this by selecting a quantile level whose overestimation rate on a calibration set meets a user-specified budget. It reports a global overestimation rate of 0.349 against a budget of 0.35, and a risk-pass on all three of its datasets.

This paper asks whether that guarantee survives contact with the conditions a deployed terminal actually operates in, and what the remedies cost. We do not propose a method. Both mechanisms we evaluate are published, and Section~\ref{sec:prior} is explicit about which paper each comes from.

\subsection{What is new in this paper}
\label{sec:new}

Novelty in a measurement paper is novelty of \emph{knowledge}, not of algorithm. Each of the following is, to our knowledge, unreported, and each is a fact about published methods or about LEO links that was not previously on record. We searched for prior work on each before claiming it; Section~\ref{sec:prior} gives the full prior-art accounting.

\begin{enumerate}
\item \textbf{BG-CFQS cannot serve a risk budget below the low end of its candidate quantile set.} The achieved overestimation rate is identical at budgets of 0.05, 0.10 and 0.15 on all four datasets, overshooting by up to 3.7x, because the selection interval collapses and the method returns the same quantile. Derived from their Algorithm~1, confirmed against their Table~II, and measured. Their own evaluation is conducted at a single budget where the floor never binds, so it cannot surface. This is the strongest result in the paper, and a reader can check it in ten minutes. (Section~\ref{sec:floor})

\item \textbf{The method's risk guarantee depends on an exchangeable calibration split.} We recover every published quantity under a random split, including the risk-pass column at 3/3, and 1/3 under a contiguous temporal split. That split conformal degrades under non-exchangeability is textbook~\cite{barber2023beyond}; that \emph{this method, on these traces, moves from 3/3 to 1/3} is not on record, and it is the difference between the method working and not working in deployment. (Section~\ref{sec:split})

\item \textbf{Measured serving-satellite geometry predicts throughput level but not residual structure.} StarNet~\cite{liu2025starnet} establishes that elevation, distance and visible-satellite count correlate with throughput, and we reproduce that. We show they carry no information about the forecaster's \emph{uncertainty}: conditioning a calibration layer on them is worse than not conditioning at all, on all three of the StarNet traces. The level/residual distinction is not drawn for these features anywhere we could find, and it is what determines whether the covariates are useful for risk control. (Section~\ref{sec:groups})

\item \textbf{Static calibration misses a risk budget in both directions on LEO links, with the sign set by the drift.} It overshoots on one dataset and undershoots by a fifth on another. The undershoot half is under-reported in a literature that emphasises under-coverage, and it is not a safe failure: it silently withholds capacity the operator said it would risk. (Section~\ref{sec:online})

\item \textbf{A negative result on predicting when group conditioning helps.} The intuition that measurable heterogeneity predicts benefit is appealing and cheap to implement. It is wrong on our data, in two independent ways, and we report the failure with the mechanism. (Section~\ref{sec:gate})

\item \textbf{Static conformal calibration cannot improve a ranking, only an operating point.} A static bound is the point forecast plus a constant; adding a constant does not reorder predictions; AUPRC depends only on the order. We observe the consequence exactly, to three decimal places, on all three locations: point forecast and static bound score identical AUPRC for latency spike detection. An online per-regime bound is not a constant offset, so it can reorder, and improves AUPRC by 43\% and 51\% on two of three links. This is a reason to prefer online calibration that is independent of every risk result in this paper. (Section~\ref{sec:latency})

\item \textbf{A price-invariant decision rule for whether risk control is worth deploying.} Whether a conservative allocator beats an aggressive one depends on prices, and under commodity pricing it does not. The invariant is the break-even ratio: how much more a violated session-hour must cost than a sold one. It lands at 3.87x, 3.92x and 4.05x on three independent links, within 5\% of each other. Consumer broadband does not clear it; enterprise and URLLC services do. (Section~\ref{sec:cost})
\end{enumerate}

\paragraph{What we do not claim.} The two mechanisms evaluated here are both published: group-conditional online recalibration is GCACI~\cite{ramalingam2025noregret}, and deciding whether to condition from data is Clustered Conformal Prediction~\cite{ding2023clustered}. Our implementation of the former was written independently and is its naive special case. We include both as baselines rather than presenting either as ours.

\section{Setup}
\label{sec:setup}

\paragraph{Datasets.} Three Starlink traces released with StarNet~\cite{liu2025starnet}, covering Chicago (1,123,832 samples, 2024-04-26 to 05-28), Osnabruck (613,295, 2024-07-13 to 07-31) and Victoria (145,053, 2024-07-11 to 07-28), each at 1\,Hz with serving-satellite identity, elevation, distance, visible-satellite count and co-located weather. Plus the per-second iperf release of WetLinks~\cite{laniewski2024wetlinks}, 1,019,109 samples of measured capacity at Osnabruck over 180 days, which is an independent measurement campaign with a different instrument and no satellite telemetry.

\paragraph{Forecaster.} The StarNet backbone: a two-layer GRU sequence-to-sequence model with a periodical embedding over the 15-second cycle and attention. We reproduce its published accuracy within 1.1\% RMSE and 2.4\% MAE on average across the three locations (Appendix~\ref{app:gate}), which is the gate everything downstream rests on.

\paragraph{Metrics.} Following BG-CFQS exactly, so the numbers are comparable: OverRate, the fraction of decisions where the bound exceeded the realised capacity; MPE, the mean positive error; P95+Err. Reported globally and on the two risk slices the paper defines, \textbf{P30} and \textbf{P10}, the lowest 30\% and 10\% of realised throughput. These are where over-allocation actually drops sessions.

\paragraph{Splits.} Contiguous temporal blocks, 60/20/20, with a purge band so no time step crosses a boundary. Every run carries a leak check.

\paragraph{Use of generative AI.} Generative AI tools were used to assist with implementing the code and with drafting and editing the text. Every number reported here was produced by the authors' code and read from saved result files, and the authors checked all content and take responsibility for it.

\section{What a risk budget is supposed to buy}
\label{sec:budget}

An allocator sets a budget $\varepsilon$ and expects the bound to over-promise at most that often. The whole value proposition is that $\varepsilon$ is a dial the operator sets, not a number the method reports afterwards.

Two things can go wrong, and both do:
\begin{itemize}
\item the achieved rate can \textbf{exceed} the budget, which drops sessions;
\item the achieved rate can fall \textbf{below} the budget, which wastes capacity the operator explicitly said it was willing to risk.
\end{itemize}

The second failure is under-discussed. It is not safe, it is expensive: an allocator that spends 0.29 of a 0.35 budget is leaving utilisation on the table and getting no risk reduction the operator asked for.

\section{BG-CFQS}

\subsection{The guarantee is conditional on the split}
\label{sec:split}

Reproducing BG-CFQS with its published configuration ($L=75$, $H=15$, $\varepsilon=0.35$, $T=[0.15, 0.40]$, $\delta=0.05$, $M=5$, XGBoost with pinball loss) under contiguous temporal splits, the conditional failure it reports comes out clearly, but its risk column does not: average OverRate 0.391 against their 0.349, risk pass 0/3 against 3/3, while accuracy came out \emph{better} than published. Better accuracy with worse risk is not a signature of a broken reimplementation, so we looked for the confound. Table~\ref{tab:split} holds everything constant and varies only the split.

\begin{table}[t]
\caption{BG-CFQS under the two splits, averaged over the three StarNet traces. Only the split differs between the first two rows.}
\label{tab:split}
\begin{tabular}{lcccc}
\toprule
Split & OverRate & P30 & P10 & Risk pass \\
\midrule
Temporal & 0.377 & 0.750 & 0.909 & 1/3 \\
Random (exchangeable) & 0.340 & 0.671 & 0.848 & 3/3 \\
Published~\cite{xie2026bgcfqs} & 0.349 & 0.65 to 0.71 & 0.83 to 0.86 & 3/3 \\
\bottomrule
\end{tabular}
\end{table}

Under the exchangeable split every published quantity lands in range. The reimplementation is correct, and the guarantee holds under the assumption it inherits from split conformal prediction~\cite{vovk2005algorithmic}.

It does not survive a temporal split. This is not a criticism of the paper, which does not report the temporal case, and it is not a discovery about conformal prediction, where non-exchangeability is textbook~\cite{barber2023beyond}. It is a statement about deployment: the assumption is not available to a terminal, and the guarantee is the reason to use the method.

The detail that matters most is in the P10 column of the random split. At 0.848, in the arm where the method passes its budget 3/3, the conditional failure is nearly as bad as under drift. So the tail failure is intrinsic to selecting one global quantile and is not a symptom of non-stationarity. The two problems are independent, and a fix for one is not a fix for the other.

\subsection{Budgets below 0.15 are unservable}
\label{sec:floor}

This is a structural limitation and it holds under any split. From Algorithm~1 of the paper, verbatim:

\begin{verbatim}
4:  if R(tau_max) <= e then
5:      Set boundary interval [t-, t+] = [tau_max, tau_max]
6:  else if R(tau_min) > e then
7:      Set boundary interval [t-, t+] = [tau_min, tau_min]
...
21: Build Q_fine = LinSpace(t-, t+, M)
23: if exists tau in Q_fine such that R(tau) <= e then
24:     Select tau* = argmin A(tau) over the feasible set
25: else
26:     Select tau* = argmin A(tau) + lambda max(R(tau) - e, 0)
\end{verbatim}

When the requested budget is tighter than the risk achievable at the most conservative candidate $\tau_{\min}$, line~6 fires. The interval collapses to a single point, line~21 produces $M$ copies of $\tau_{\min}$, the feasible set at line~23 is empty, and line~26 returns $\tau_{\min}$ however small the budget gets.

\begin{figure}[t]
\centering
\includegraphics[width=\linewidth]{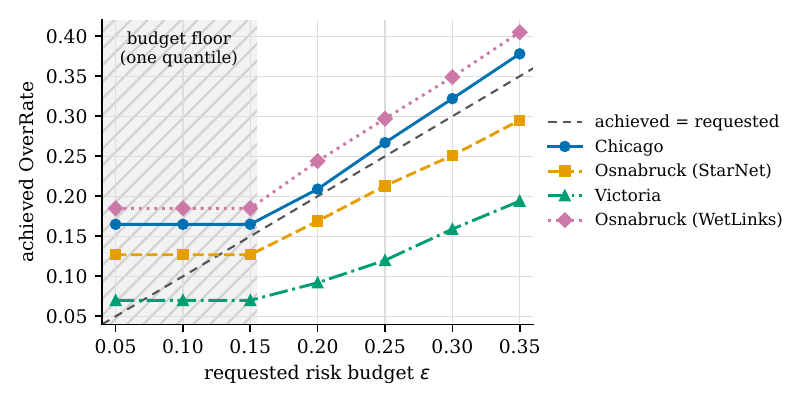}
\caption{BG-CFQS achieved OverRate against the requested budget. Every line is flat across the shaded band of budgets at or below 0.15, because the method returns the same quantile there. The dashed line marks achieved equal to requested.}
\Description{A line chart with four datasets. Each line is flat for requested budgets 0.05, 0.10 and 0.15, then rises with the budget. At a requested budget of 0.05, every line sits above the dashed target line.}
\label{fig:floor}
\end{figure}

With their published $T = [0.15, 0.40]$ (their Table~II), the achieved rate therefore pins at $R(0.15)$. The signature is an achieved rate that is \emph{identical} at every budget at or below 0.15, and it appears on all four datasets (Table~\ref{tab:floor}, Figure~\ref{fig:floor}).

\begin{table}[t]
\caption{BG-CFQS achieved OverRate across the budget sweep. The three tightest budgets return the same number on every dataset.}
\label{tab:floor}
\begin{tabular}{lcccccccc}
\toprule
Budget & 0.05 & 0.10 & 0.15 & 0.20 & 0.25 & 0.30 & 0.35 & Overshoot at 0.05 \\
\midrule
Chicago & 0.165 & 0.165 & 0.165 & 0.209 & 0.267 & 0.322 & 0.378 & 3.3x \\
Osnabruck (StarNet) & 0.127 & 0.127 & 0.127 & 0.169 & 0.213 & 0.251 & 0.295 & 2.5x \\
Victoria & 0.070 & 0.070 & 0.070 & 0.092 & 0.120 & 0.159 & 0.194 & 1.4x \\
Osnabruck (WetLinks) & 0.185 & 0.185 & 0.185 & 0.244 & 0.297 & 0.349 & 0.405 & 3.7x \\
\bottomrule
\end{tabular}
\end{table}

The floor value differs by dataset because it is $R(\tau_{\min})$, the risk the most conservative available quantile happens to achieve on that link. The flatness does not: on every dataset the three tightest budgets return the same number, because the method is returning the same quantile. For comparison, online recalibration on the same traces tracks every one of those budgets within 1\% (Section~\ref{sec:online}).

Their evaluation is conducted at $\varepsilon = 0.35$ (their Table~II, ``Default risk budget''), where the floor never binds on any of the four datasets, so their own results cannot expose it.

A budget of 0.35 permits over-promising on more than a third of decisions. We know of no allocator that would accept that, which makes the tighter budgets the interesting ones and the floor a practical obstacle rather than a curiosity.

Widening $T$ fixes it, and would no longer be their method. The general lesson is about the method class: budget-guided selection over a bounded candidate set inherits that set's bounds as hard limits on the achievable risk.

\section{The remedy, and how much of it works}

\subsection{Two mechanisms}

The natural fix has two parts, and they are separable:
\begin{itemize}
\item \textbf{Online recalibration.} Stop treating the operating point as a constant. Update it from outcomes as they are observed. This is ACI~\cite{gibbs2021adaptive}.
\item \textbf{Group conditioning.} Maintain a separate operating point per regime, where a regime is a bucket over covariates the terminal can compute at prediction time. Combined with the above this is GCACI~\cite{ramalingam2025noregret}.
\end{itemize}

\subsection{The online mechanism replicates}
\label{sec:online}

\begin{figure}[t]
\centering
\includegraphics[width=0.78\linewidth]{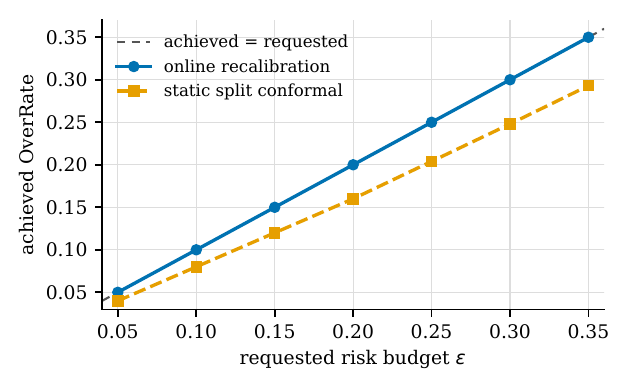}
\caption{Budget tracking on the StarNet traces, mean of three locations. Online recalibration sits on the dashed target line at every budget; static split conformal runs 16 to 21\% below it.}
\Description{A line chart of achieved against requested risk budget. The online line lies on the dashed diagonal at all seven budgets. The static line lies below it at every budget, by a gap that grows from 0.010 to 0.057.}
\label{fig:tracking}
\end{figure}

Table~\ref{tab:tracking} and Figure~\ref{fig:tracking} show budget tracking across the sweep, averaged over three locations. The online layer tracks within 0.8\% everywhere. Static split conformal is 16 to 21\% \textbf{low}, the second failure mode from Section~\ref{sec:budget}.

\begin{table}[t]
\caption{Requested budget against achieved OverRate, mean of the three StarNet locations.}
\label{tab:tracking}
\begin{tabular}{lccccccc}
\toprule
Budget & 0.05 & 0.10 & 0.15 & 0.20 & 0.25 & 0.30 & 0.35 \\
\midrule
Online achieved & 0.050 & 0.100 & 0.150 & 0.200 & 0.250 & 0.300 & 0.350 \\
Static achieved & 0.040 & 0.080 & 0.120 & 0.160 & 0.204 & 0.248 & 0.293 \\
Static error & $-21\%$ & $-20\%$ & $-20\%$ & $-20\%$ & $-19\%$ & $-17\%$ & $-16\%$ \\
\bottomrule
\end{tabular}
\end{table}

Note the sign against the WetLinks result, where static calibration \emph{overshot} (0.423 against 0.35) because the test month was slower than the calibration month. Here the test period is easier and it undershoots. The direction of the miss is a property of the drift, not of the method, which is a stronger claim than either dataset alone supports.

Table~\ref{tab:matched} makes the comparison a naive table gets wrong: online against static, at matched achieved risk.

\begin{table}[t]
\caption{Online against static per-regime calibration at matched achieved risk, mean of the three StarNet locations. The 0.35 row is omitted because static calibration never reaches 0.35.}
\label{tab:matched}
\begin{tabular}{cccccc}
\toprule
Achieved risk & Online P10 & Static P10 & Change & Online MAE & Static MAE \\
\midrule
0.050 & 0.145 & 0.170 & $-14.6\%$ & 51.54 & 52.66 \\
0.100 & 0.256 & 0.289 & $-11.3\%$ & 39.62 & 40.64 \\
0.150 & 0.357 & 0.398 & $-10.4\%$ & 33.67 & 34.47 \\
0.200 & 0.438 & 0.499 & $-12.1\%$ & 29.76 & 30.53 \\
0.250 & 0.521 & 0.589 & $-11.6\%$ & 27.23 & 28.00 \\
0.300 & 0.598 & 0.666 & $-10.1\%$ & 25.36 & 26.39 \\
\bottomrule
\end{tabular}
\end{table}

Online calibration has better tail risk and better accuracy at every operating point. \textbf{Methods must be compared at matched achieved risk}: ranked by raw P10 alone, static BG-CFQS appears best, but it is spending 0.289 of a 0.35 budget and conservatism buys P10 directly.

\subsection{The group-conditional mechanism mostly does not}
\label{sec:groups}

Table~\ref{tab:axes} reports online calibration at budget 0.35, mean P10 over three locations. With $n = 3{,}718$ in the Chicago P10 slice the standard error is 0.008, so these differences are real.

\begin{table}[t]
\caption{Effect of each regime set on P10 OverRate under online calibration at budget 0.35, mean of the three StarNet locations.}
\label{tab:axes}
\begin{tabular}{lcc}
\toprule
Regime axes & P10 & Change against no conditioning \\
\midrule
None & 0.666 & n/a \\
Level (observed look-back mean) & 0.673 & $+1.1\%$ \\
Visible satellite count & 0.676 & $+1.4\%$ \\
Serving elevation & 0.685 & $+2.9\%$ \\
All measured geometry & 0.693 & $+4.0\%$ \\
15\,s scheduling phase & 0.694 & $+4.2\%$ \\
All four axes & 0.709 & $+6.3\%$ \\
\bottomrule
\end{tabular}
\end{table}

Every axis is worse than not conditioning, including the \emph{measured} serving-satellite geometry. StarNet reports that elevation, distance and visible count correlate with throughput \emph{level}, and we reproduce that. They do not carry \emph{residual} structure, which is what a calibration layer needs, and the distinction matters.

The result is not an artifact of the learning rate. Across seven values spanning two orders of magnitude, conditioning beat no conditioning in 8 of 42 cells, and seven of the eight are Victoria. On Chicago it loses at every setting.

\begin{figure}[t]
\centering
\includegraphics[width=0.78\linewidth]{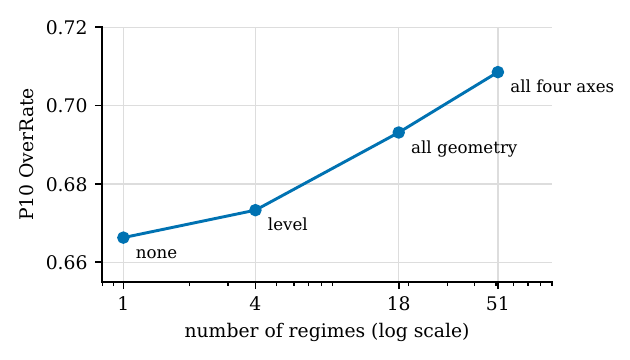}
\caption{P10 OverRate against the number of regimes, online calibration at budget 0.35, mean of the three StarNet locations. Finer partitions are worse.}
\Description{A line chart with four points on a logarithmic horizontal axis: one regime at 0.666, four regimes at 0.673, eighteen regimes at 0.693 and fifty-one regimes at 0.709. The line rises steadily.}
\label{fig:regimes}
\end{figure}

The degradation tracks regime count rather than axis choice (Figure~\ref{fig:regimes}): 1 regime 0.666, 4 regimes 0.673, 18 regimes 0.693, 51 regimes 0.709, with a correlation of $-0.571$ between $\log_{10}$(outcomes per regime) and P10. Online per-regime calibration must converge one parameter per regime from a \emph{stream}, and 51 regimes leave about 400 outcomes each. Static per-regime calibration, which fits once, is less affected, and conditioning on level still helps it by about 3\%. This is the estimation-noise effect \citet{ding2023clustered} describe for class-conditional conformal prediction, appearing here in an online setting.

\subsection{A negative result on predicting when conditioning helps}
\label{sec:gate}

Because conditioning helps on one dataset and hurts on three, we tried to build a test that decides which case a trace is in, using only the calibration split.

The statistic: fit one offset per regime on the calibration split, weight each by the standard error of an empirical quantile~\cite{siddiqui1960distribution}, and run Cochran's $Q$~\cite{cochran1954combination} against $\chi^2$, requiring both significance and a substantial $I^2$~\cite{higgins2002quantifying}. This is the standard heterogeneity test from meta-analysis, and doing it for conformal calibration is Clustered Conformal Prediction~\cite{ding2023clustered} in binary form.

It does not work, for two independent reasons.

It fires everywhere. On Chicago, where conditioning costs 6.1\%, the test returns $Q = 18.3$ on 3 degrees of freedom, $p = 0.0004$, $I^2 = 0.84$, for an offset spread of 1.99\,Mbps. With about 1,200 calibration points per regime a 2\,Mbps difference is comfortably significant, and $I^2$ is scale free so it agrees that the difference is real. Both hurdles ask whether a difference \emph{exists}; neither asks whether it is worth anything.

And the statistic does not order the outcome (Table~\ref{tab:gate}). Osnabruck has the largest pre-test spread and gets no benefit; Victoria has a smaller one and gets the only benefit.

\begin{table}[t]
\caption{Offset spread measured on the calibration split, against the measured effect of conditioning on P10.}
\label{tab:gate}
\begin{tabular}{lcc}
\toprule
Location & Pre-test spread & Measured effect \\
\midrule
Chicago & 1.99 Mbps & $+6.1\%$ (hurts) \\
Osnabruck & 5.83 Mbps & $+0.1\%$ (neutral) \\
Victoria & 5.17 Mbps & $-1.7\%$ (helps) \\
\bottomrule
\end{tabular}
\end{table}

\paragraph{A correction to our own earlier analysis.} We initially recorded this relationship as monotonic across four datasets. Those figures were taken from the offsets \emph{applied during the test replay}, a post-hoc quantity not available before a decision is made. Computed correctly on the calibration split, the relationship does not hold. We report this because the failure is the result: the intuition that heterogeneity predicts benefit is appealing, cheap to implement, and wrong on our data.

What would be needed is a \textbf{stability} test rather than a spread test. Both Osnabruck and Victoria have real per-regime differences on the calibration split; only Victoria's persist into the test period. That is a different question and we did not attempt it.

\subsection{Static calibration cannot improve a ranking}
\label{sec:latency}

The same layer serves latency spike detection with the bound direction reversed: an upper bound on latency, where the risk event is underestimation. Following \citet{casparsen2026latency}, each 15-second period aligned to the recovered phase is Good if its latency stays within 50\,ms and Degraded otherwise. At 1\,Hz a period holds fifteen samples, so their 99\% criterion is effectively a maximum.

The two-directional failure of static calibration reappears, more severely than for throughput. At a budget of 0.10 the online layer achieves 0.101, 0.100 and 0.088 on Chicago, Osnabruck and Victoria. Static calibration achieves 0.143 ($+43\%$), 0.313 ($+213\%$) and 0.033 ($-67\%$). That the same failure appears with a different target and the bound direction reversed is stronger evidence that it is a property of the link than the throughput result alone.

\begin{table}[t]
\caption{Latency spike detection, AUPRC with the multiple of chance in brackets. The point forecast and the static bound are identical on every link.}
\label{tab:auprc}
\begin{tabular}{lcccc}
\toprule
Location & Chance & Point forecast & Static bound & Online bound \\
\midrule
Chicago & 0.046 & 0.628 (13.6x) & 0.628 (13.6x) & 0.600 (13.0x) \\
Osnabruck & 0.065 & 0.348 (5.4x) & 0.348 (5.4x) & 0.527 (8.1x) \\
Victoria & 0.123 & 0.360 (2.9x) & 0.360 (2.9x) & 0.516 (4.2x) \\
\bottomrule
\end{tabular}
\end{table}

Table~\ref{tab:auprc} shows the ranking result. The point forecast and the static bound have identical AUPRC on all three locations, and this is a necessity rather than a coincidence. A static conformal bound is the point forecast plus a constant. Adding a constant cannot change the order of the predictions, and AUPRC depends only on that order. Static calibration can therefore move the operating point along a precision-recall curve but can never improve the curve. The online per-regime bound applies different corrections in different regimes and at different times, so it can reorder predictions, and on two of the three links it improves AUPRC by 51\% and 43\%. On Chicago it is slightly worse. F1 is less favourable to the online method: it wins on Osnabruck (0.429 against 0.349) and Victoria (0.348 against 0.229) and loses on Chicago (0.508 against 0.601), where higher recall costs more precision than it gains.

\subsection{A price-invariant rule for deploying risk control}
\label{sec:cost}

Admission control admits $\lfloor b/10 \rfloor$ sessions of 10\,Mbps against a bound $b$, following BG-CFQS. Availability is delivered over promised service time. Table~\ref{tab:avail} shows that calibration moves each link from roughly one nine to two, cuts the number of outages by five to six times, and shortens the mean outage from about nine seconds to about five. No policy reaches three nines; the best is 2.13.

\begin{table}[t]
\caption{Service availability by policy on the three StarNet links.}
\label{tab:avail}
\begin{tabular}{llcccc}
\toprule
Location & Policy & Availability & Nines & Outages & Mean time to repair \\
\midrule
Chicago & Point forecast & 93.50\% & 1.19 & 8,967 & 9.0 s \\
Chicago & Online, budget 0.05 & 99.16\% & 2.08 & 1,494 & 5.3 s \\
Osnabruck & Point forecast & 93.37\% & 1.18 & 4,514 & 9.2 s \\
Osnabruck & Static, budget 0.05 & 99.18\% & 2.09 & 651 & 5.1 s \\
Victoria & Point forecast & 93.44\% & 1.18 & 1,001 & 8.8 s \\
Victoria & Static, budget 0.05 & 99.25\% & 2.13 & 146 & 5.0 s \\
\bottomrule
\end{tabular}
\end{table}

Whether this is worth buying depends on prices. The cost model combines tiered SLA credits for over-allocation with foregone revenue for under-allocation, which is what stops the cheapest policy from being ``admit nothing''. Under our default illustrative prices the aggressive policy is cheapest: on Chicago, 0.473 per hour at budget 0.35 against 0.629 at budget 0.05. That is a statement about the prices, not about the method. Foregone revenue accrues on every unsold session-hour, while SLA credits are a fraction of a small fee and only bite below three nines.

The quantity that does not depend on prices is the break-even ratio: how much more a violated session-hour must cost than a sold one before the risk-controlled policy pays for itself. It is 3.87x on Chicago, 3.92x on Osnabruck and 4.05x on Victoria, within 5\% across three links on two continents. Consumer broadband, where a degraded minute costs little more than the revenue on it, does not clear a ratio of 4. Enterprise connectivity and URLLC services, where credits run to 10\% to 25\% of a monthly fee, clear it comfortably. The model prices violations linearly whereas real SLA remedies are stepped and capped, so 4 is an approximation to the crossing point.

\section{What is and is not new here}
\label{sec:prior}

Stated plainly, because the answer is mostly ``not new''. Table~\ref{tab:prior} lists every mechanism used and where it comes from.

\begin{table}[t]
\caption{Every mechanism evaluated in this paper is prior work.}
\label{tab:prior}
\begin{tabular}{p{0.42\linewidth}p{0.5\linewidth}}
\toprule
Mechanism & Prior work \\
\midrule
Per-regime conformal calibration & Mondrian conformal prediction~\cite{vovk2003mondrian} \\
Per-regime \emph{online} conformal calibration & GCACI~\cite{ramalingam2025noregret}; concurrently~\cite{angelopoulos2025gradient} \\
Parameter-free version & POGO~\cite{bharti2026pogo} \\
Online control of a user-specified risk & Rolling RC~\cite{feldman2023rollingrc} \\
Deciding whether to condition, from data & Clustered Conformal Prediction~\cite{ding2023clustered}; AFCP~\cite{afcp2024} \\
Conformal risk control for resource allocation & URLLC scheduling~\cite{cohen2023urllc} \\
Admission control from a safe throughput bound & BG-CFQS~\cite{xie2026bgcfqs}, eqs.~(23) to (26) \\
\bottomrule
\end{tabular}
\end{table}

Our implementation of per-regime adaptive calibration was written independently and is the naive special case of GCACI: for a partition, GCACI's group membership vector is one-hot and its update reduces to one parameter per regime. We include GCACI and Rolling RC as baselines.

\textbf{What this paper contributes is the findings in Section~\ref{sec:new}.} They are new knowledge, not new machinery: a structural limitation in a published method that its own evaluation cannot surface, the condition its guarantee depends on, a covariate class that predicts level but not uncertainty, a two-directional failure of static calibration, a limit on what static calibration can do for detection, a price-invariant deployment rule, and a negative result.

We were careful about this distinction because we got it wrong first. The project was designed around a per-regime calibration layer believed to be novel, and that belief survived until it was checked against the literature rather than against our own reading of two papers. The check cost a day and removed the methods claim entirely. We record that here because the temptation in a project like this is to check late, or to check in a way that confirms.

\section{Limitations}

\begin{itemize}
\item \textbf{Four datasets, three of them from one measurement campaign.} Any cross-dataset claim rests on four observations.
\item \textbf{Cross-location means two European sites 150\,km apart} on the WetLinks side, and three sites from one collection on the StarNet side.
\item \textbf{The 15-sample constraint on WetLinks} caps look-back plus horizon at 15, so neither StarNet's 30/5 nor BG-CFQS's 75/15 configuration runs there.
\item \textbf{The online layer's budget control is empirical}, not a finite-sample guarantee, and it requires outcome feedback after each decision.
\item \textbf{No live terminal.} Every result is replay over recorded traces.
\item \textbf{We did not search the shrinkage and empirical-Bayes literature}, where the pooling-versus-conditioning question has its own long history. Our negative result in Section~\ref{sec:gate} should be read against that gap.
\end{itemize}

\bibliographystyle{ACM-Reference-Format}
\bibliography{references}

\appendix

\section{Reproduction gate}
\label{app:gate}

StarNet with look-back 30, output 5, their published per-location step sizes and a contiguous 8:2 split (Table~\ref{tab:gatestarnet}). Gaps scatter in both directions. We do not reproduce their whole-trace scaler, which fits before splitting; ours fits on training windows only.

\begin{table}[h]
\caption{StarNet reproduction against the published results.}
\label{tab:gatestarnet}
\begin{tabular}{lcccccc}
\toprule
Location & RMSE & Published & Gap & MAE & Published & Gap \\
\midrule
Chicago & 42.56 & 40.33 & $+5.5\%$ & 31.94 & 29.88 & $+6.9\%$ \\
Victoria & 38.38 & 41.08 & $-6.6\%$ & 29.55 & 30.84 & $-4.2\%$ \\
Osnabruck & 38.24 & 36.48 & $+4.8\%$ & 28.42 & 27.11 & $+4.8\%$ \\
\midrule
Average & 39.73 & 39.30 & $+1.1\%$ & 29.97 & 29.28 & $+2.4\%$ \\
\bottomrule
\end{tabular}
\end{table}

\section{Artifacts}
\label{app:artifacts}

All code, run configurations and result files will be released publicly on acceptance. Every table in this paper is generated from a saved result file with its configuration snapshot beside it. The two reproduction gates, the split sensitivity experiment, the learning-rate sweep and the heterogeneity gate are each a single script.

\end{document}